\documentclass[conference]{IEEEtran}
\usepackage{times}

\usepackage[numbers]{natbib}
\usepackage{multicol}
\usepackage[bookmarks=true]{hyperref}
\usepackage{graphicx}
\usepackage{amsmath}
\usepackage{cuted}
\usepackage{capt-of}
\usepackage{xcolor}
\usepackage{amssymb}
\usepackage[ruled,vlined]{algorithm2e}
\usepackage{colortbl}
\usepackage{comment}
\usepackage{balance}

\begin{document}

% paper title
\title{Damage Adaptation in Seconds \\ for Architected Materials}

\author{James Avtges\textsuperscript{*}, Jake Ketchum\textsuperscript{*}, Helena Young, Taekyoung Kim, Ryan L. Truby, Todd D. Murphey \\ Northwestern University, Evanston, IL, USA \\ \textsuperscript{*}Equal Contribution. Email: \{javtges, jketchum\}@u.northwestern.edu}

\maketitle

\begin{strip}
    \centering
    \vspace{-35pt}
    \includegraphics[width=1\linewidth]{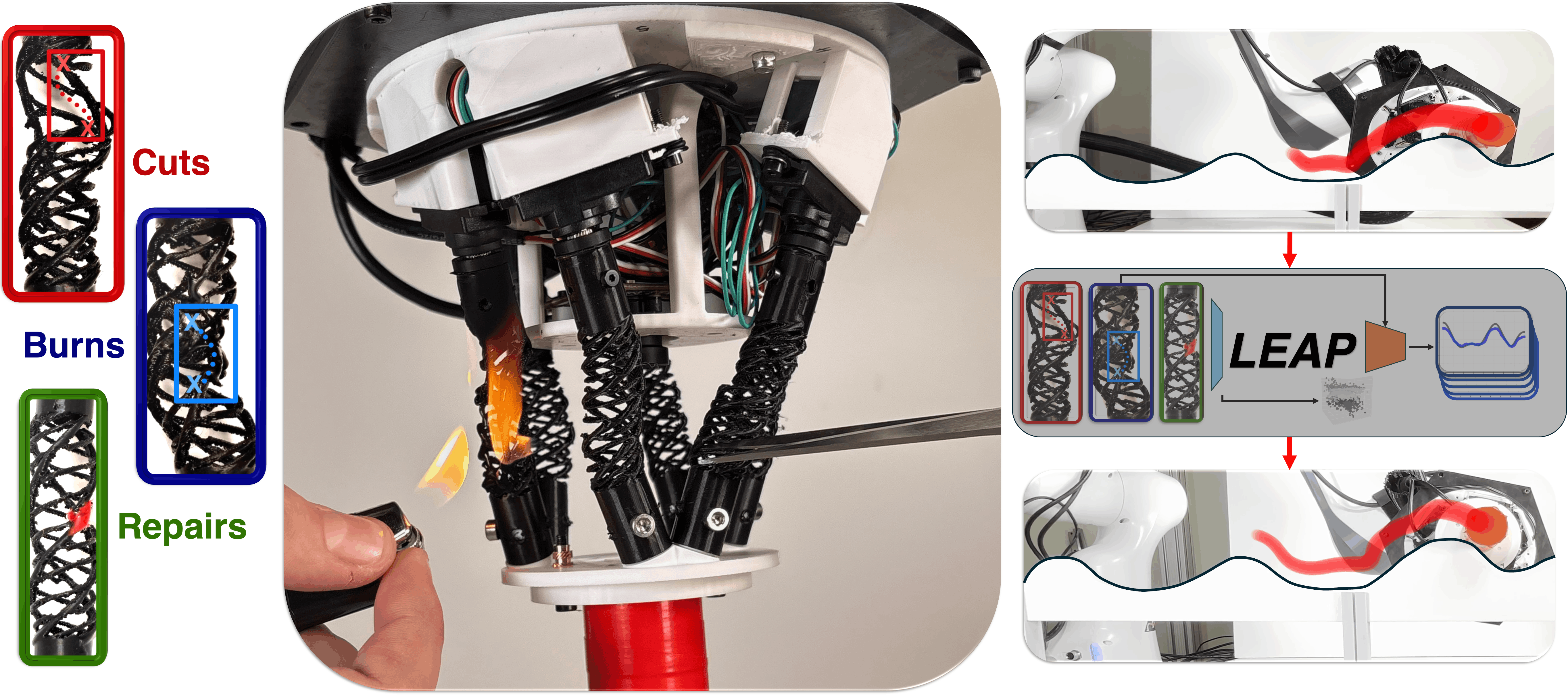}
    \captionof{figure}{\textbf{Latent Ensemble Adaptive Proprioception (LEAP):} Modeling damage in architected materials enables simple yet robust ensemble methods to adapt to damage in seconds. We evaluate our approach on a soft wrist that completes a contour tracing task using proprioception alone. The wrist is damaged by melting actuators (left), cutting with scissors (right), and repairing them with glue. After adaptation, LEAP recovers up to all of the undamaged performance.}
    \label{fig:placeholder_figure1}
\end{strip}

\begin{abstract}
Adaptation to damages and in-situ physical repairs is essential for long-term robot autonomy, yet challenging outside of narrowly defined and well-anticipated bounds. In this work we proprioceptively adapt to catastrophic damage in soft-actuated systems in under one minute. Architected materials are well equipped for adaptation: actuator failure occurs gradually rather than acutely, and damage can be described in a low-dimensional, discrete coordinate space. Surprisingly, latent damage representations plus a simple yet robust ensemble method is sufficient for adapting to unseen damage in real-time. Moreover, we identify conditions under which exponential sample complexity collapses to linear sample complexity for learned representations of architected materials, a concrete advantage over rigid components or continuum soft mechanisms. We demonstrate LEAP, our method for adaptive proprioception, via a tracing task for a 6DoF soft wrist based on Handed Shearing Auxetic (HSA) actuators. Our algorithm is able to adapt to cuts, burns, and actuator repairs, enabling simulation-free real-time adaptation that is critical for realizing the promise of soft robots outside the lab. Videos and more information are available at \url{https://murpheylab.github.io/leap}.

\end{abstract}

\IEEEpeerreviewmaketitle

\section{Introduction}

A major barrier to bringing soft robots out of the lab is the reliability of the compliant structural components, transducers, and actuators. Soft robots use mechanical deformation to perform work, causing material failure through creep and fatigue. Actuators often have a continuous operating life of hours to even minutes, in contrast to brushless DC motors whose operating lives can be multiple orders of magnitude longer. All of these challenges are exacerbated in human-centric spaces such as kitchens, workshops, and hospitals where there are hazards that may lacerate, wear, or melt flexible devices. 

Even within the nominal operating life of a soft actuator, the challenges associated with modeling and control have been well-documented \cite{gazzola_forward_2018}. In addition to the difficulties of modeling high-dimensional soft bodies, a soft actuator cannot be parametrically defined by constant terms that engineers typically have access to, as is the case with rigidly defined systems. Material properties, and therefore actuator models, are often hysteretic and vary with the robot's configuration, payload, and temperature. Furthermore, actuator manufacturing is often done by hand and even when produced in parallel (i.e., on the same build plate of a 3D printer), current processes produce significant actuator-to-actuator variability \cite{oh_architected_2026}.

Existing works in this field typically produce outputs that implicitly assume ``nominal'' robot conditions: either immediately after the actuator is constructed or after some defined break-in procedure. As soon as a soft robot performs significant work, however, the robot's constituent components inevitably change, causing a non-adaptive controller to be out-of-distribution with respect to its underlying model. Moreover, soft actuators often play a multifunctional role in an engineered system, providing structural integrity, motion, and contact surfaces, all of which are impacted upon experiencing damage.
\color{black}

Architected materials are advantageously positioned for damage adaptation. An architected material is a material whose structure and microscale properties are intentionally designed to produce desired functionalities and macroscale properties beyond material properties alone \cite{oh_architected_2026}. A hallmark of most architected materials is a lattice- or unit cell-based structure, such as the structures found in auxetics, kirigami, and active lattices. This provides a discrete mechanism to parameterize material damage, whereas soft body structures and rigid-linked systems have continuous parameterizations of structure damage. For example, a recent method to adapt to broken legs in a rigid quadruped \cite{omnibodied_skild} requires robust modeling of thousands of different damage states. The compliant nature of architected materials provides degeneracies---within each lattice member all damages are equivalent.

Since architected materials can replace multiple components in a traditional robotics setting such as gears, tendons, hinges, and joints, the multifunctional lattice of the material also provides redundancies, whereas the constituent components of a rigid analog are all single points of failure.

Furthermore, failure modes in architected materials are often non-acute. The materials themselves can continue operation, albeit with changes to the overall properties of the mechanism, before catastrophic material failure \cite{carton_bridging_2025}. Architected materials in particular utilize redundancies in their discrete structure such that functionality can remain despite multiple links being severed.

In this work we introduce Latent Ensemble Adaptive Proprioception (LEAP). Our method utilizes these inherent advantages of architected materials to adapt to fully-unseen actuator damage in seconds on hardware. A key insight of our work is parameterizing the structure of architected materials into discrete coordinates for the purpose of latent modeling. This allows characterizing the damage state using a representation that is, importantly, independent of the robot state itself. While our method does not adapt to all types of damage in a single predictive step, we use this embedding as a means of adapting accurately and in near real-time. By modeling the meta-state of soft actuators as latent variables, ``adaptation" and ``damage" can be viewed both as functions of the same latent dynamics.

The contributions of this work are:
\begin{enumerate}
    % \item We provide a new way of evaluating robots by characterizing their responses to changes in latent dynamics: via wear, breakages, and in-situ repairs.
    \item LEAP, a method for using an ensemble of meta-state encodings for damage adaptation in architected materials. In 6-DoF force/torque proprioception, LEAP requires only seconds of data and achieves a 90\% reduction in proprioceptive force/torque prediction loss.
    \item A novel 6-DoF soft wrist capable of proprioceptive force/torque sensing and rapidly adapting to changing dynamics.
    \item We identify properties of architected materials that provide unique advantages for adaptation over rigid and continuum soft robots.
\end{enumerate}

\section{On the Repair of Soft Robots}

Well-defined calibration procedures in response to a motor or sensor repair have been essential elements of rigid mechanical systems for decades \cite{roth_overview_1987}. However, for soft systems, the concept of ``repairing'' a compliant actuator is typically overlooked, despite most soft robots being assembled by hand---and the ability to easily do so being one of their most transformative characteristics.

Physical repair is a vital prerequisite for the practical deployment of soft robots at scale: soft mechanisms function at a low enough degree of kinematic precision and net force output that many tools and methods used by hobbyists work well for repairing architected materials. Furthermore, repair of flexible, strain dependent actuators enables unique opportunities for in-situ mechanical adaptation when using direct replacements may be infeasible or impossible. Most repairs to flexible polymers meaningfully change the physical properties of the device; despite this, repair is generally not included in the class of settings where adaptation is necessary. 

By representing the state of a damaged actuator entirely in a latent space, our method makes no distinctions between what is a ``damage'' and what is a ``repair" to either the 6-DoF wrist or its constituent HSAs. Our method provides a new context for understanding repair---and generally the assembly of high variability components---in the fields of soft robotics and architected materials as an intermediate step requiring adaptation. Our method is faster and more computationally inexpensive than many traditional calibration procedures on servomotors, CNC spindles, and collaborative robot arms \cite{camera_cal, franzese2025mukcaaccurateaffordablecobot, 9702246}. 
\color{black}

\section{Related Works}

\subsection{Architected Materials}

Architected materials are characterized by macroscale properties that are driven by the geometry of their internal structure, and which vary from the bulk material properties of their constituent materials. Architected materials can exhibit exotic behaviors such as negative Poisson's ratios \cite{HSAScience}, very high strength/weight ratios, programmable responses\cite{liu2024}, and multi-stability \cite{yi2018}. Architected materials are often built around a repeating unit cell, and come in a wide variety of types including kirigami, origami, auxetic, and lattice structures, among others \cite{oh_architected_2026}. Advances in additive manufacturing have helped architected materials become promising candidates for transmission and structural elements in soft systems \cite{oh_architected_2026, Kim2024, rafsanjani2018}. 

In this work, we are concerned with lattice-type architected materials, and specifically Handed Shearing Auxetics (HSAs). HSAs are a type of rotary-to-linear soft-transmission first described in \cite{HSAScience}. They are composed of a 2D auxetic lattice, wrapped into a cylinder. The resulting structure will extend axially and diametrically when the two ends are rotated in one relative direction, and will correspondingly contract when the two ends are rotated in the other relative direction. HSAs can be manufactured from a variety of materials including steel \cite{HSAScience}, photopolymers \cite{Truby2021AAuxetics}, thermoplastics like TPU \cite{Kim2024}, or, as in this work, polypropylene. 

Their high force, fast response time, and ease of control has made HSAs a promising transmission element for soft robots. HSAs have been used in a wide variety of soft robots including quadrupeds \cite{Ketchum2023, Truby2021AAuxetics, kim2025}, parallel mechanisms \cite{11020875}, grippers \cite{chin_simple_2019}, ball-kicking robots \cite{kim2026}, and hoppers \cite{sullivan2025springbrakehandedshearingauxetics}. HSAs also have high mechanical reliability, making them a good candidate for real-world machine learning experiments \cite{avtges_real-time_2025, pmlr-v305-zhang25d} which may involve prolonged actuation during training.

\subsection{Soft Robot Proprioception}

In biology, proprioception is defined as the subconscious sensation of one's own motion, position, and force \cite{tuthill_proprioception_2018}. In robotics, proprioception refers to the ability of the robot to sense its own configuration and the forces it is experiencing. Proprioception is a key enabler of dexterity in robotic systems \cite{wang2018}. This is particularly important for soft systems, which often exhibit time-varying dynamics and manufacturing variation that make precise open-loop control challenging \cite{wang2018}. As a result, significant work has been done on sensorizing soft actuators using pneumatic strain gauges \cite{yuen2018}, fluidic innervation \cite{truby2022fluidic}, embedded magnetic elements \cite {sundaram2023}, and direct resistance measurement \cite{zhao2021}. Many of these methods have delivered promising results for single actuators, or small arrangements of actuators in a lab-bench context. However, in practice, methods like fluidic innervation that directly modify the actuators can greatly increase the cost, complexity, and fragility of engineered systems. Camera based approaches \cite{9866780} bypass many of the practical drawbacks to more invasive methods and serve to move the sensing mechanisms away from contact surfaces, which may be subject to wear or damage.

The same viscoelastic, non-linear, and time-varying response that makes open-loop control of soft actuators challenging also complicates analytical approaches to proprioceptive sensing \cite{wang2018}. Machine learning, which can discover the required relationships from data, offers one solution to managing the complexity \cite{9866780, zhu_forces_2025}. Combined with camera-based sensing, learned proprioception promises to deliver robust and performant soft proprioception. However, at present, even learned proprioception systems generally lack mechanisms for handling damage or repair, and often sense forces in a maximum of two dimensions \cite{9866780, zhu_forces_2025}. By implementing our method for proprioception into a 6-DoF soft wrist, we are able to reliably sense 6-axis forces and torques at resolutions of up to 0.1 N in the $F_x$ and $F_y$ directions.

\subsection{Damage Adaptation}

The ability to detect and adapt to damage is essential for systems designed to operate in extreme environments, in critical support roles, or in areas where human support is infeasible. This capability is particularly critical for soft mechanisms, which may be more vulnerable to damage and have more limited operating lives relative to their classical counterparts \cite{https://doi.org/10.1002/adrr.202500172, Kim2024, https://doi.org/10.1002/aisy.202400414}. Fortunately, soft mechanisms lend themselves to gradual degradation, which facilitates learning \cite{doi:10.1126/scirobotics.aav1488, bai_autonomous_2022}.  The majority of the damage adaptation canon to date can be split into two broad categories: Self modeling, in which the robot uses interactions with the environment to determine a model of its own dynamics, and control-adaptation, in which the robot uses interactions with the environment to determine an effective policy without first modeling its own configuration. 

Control-adaptation methods, which directly modify the robot policy based on dynamics changes, can provide excellent adaptation performance for well-defined tasks. In a mobility context, simulation-based search based methods have demonstrated real-world adaptation for damaged robots \cite{Cully:2015:10.1038/nature14422, chatzilygeroudis2018reset}. With quality simulation, control-adaptation methods can also handle very large dynamics changes which require new mobility strategies entirely \cite{Kriegman_2019}. However, control-adaptation methods do not directly model the system, and are thus unsuitable for use in proprioception. 

In this work, we are concerned with self-modeling methods. Self-modeling, an extension of classical system identification, involves the robot building and maintaining a representation of its own dynamics. This representation can then be integrated into a variety of sensing and control methods. Self-modeling for real-time adaptation is characterized by relatively high sample and compute efficiency. Self-modeling has proven to be a reliable and effective technique in classical robotics, where it has enabled robust adaptation to catastrophic damage states in legged robots \cite{1308802, bongard2006} and serial manipulators \cite{kwiatkowski2019}.

Many self-modeling methods function on the robot's state-space \cite{8972568} or on a lower-dimensional representation of the robot's geometry \cite{yu2025lytimetrobustinterpretablestatevariable}. However, advances in machine learning and image processing have enabled the use of visual sensing for continuous self-monitoring and adaptation \cite{chen2022fully, pokhrel2026zeroshotadaptationrobotstructural}. Self-modeling is sample intensive, but can handle very large dynamics changes, in contrast with control-adaptation methods which often implicitly rely on similarity between a pre-damage simulation and post-damage robot dynamics \cite{Cully:2015:10.1038/nature14422, chatzilygeroudis2018reset}. Where feasible, self-modeling methods are attractive because they enable the use of general purpose methods for control, guidance, and navigation.

\begin{figure*}
    \centering
    \includegraphics[width=1\linewidth]{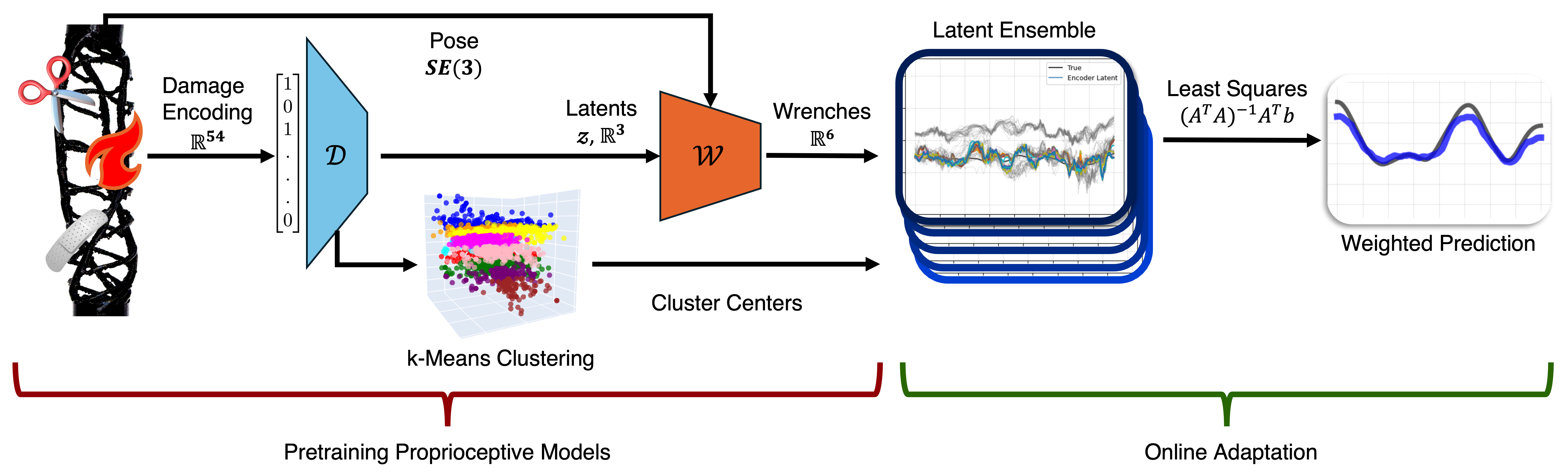}
    \caption{\textbf{Architecture of LEAP:} Our proposed approach encodes damage states into a latent representation and uses the latent structure to produce ensembles---where online data collection and regression of ensemble weightings occurs within seconds. Our architecture contains two neural networks---the first produces latent encodings of the actuator damage state. The second uses the latent variables alongside the measured displacement of the actuator in $SE(3)$ to predict the 6-dimensional forces and torques at the actuator base. After training, the latent space provides clusters from which we produce an ensemble representing a sampling of possible damage states. In inference, we learn linear weightings per-component for the ensemble via least squares which provides adaptive force/torque predictions.}
    \label{fig:placeholder3}
\end{figure*}

\section{Methods}

Our method, Latent Ensemble Adaptive Proprioception (LEAP), utilizes ensembles of damaged architected materials that are parameterized by clustered latent states. By training a damage encoder and producing the ensemble members as a pre-training process, adaptation to fully-unseen types of damage and manufacturing variations in HSAs can be achieved in seconds.

\subsection{Representing Damage in Discrete Coordinates}
\label{damage_as_coords}

The structure of HSAs and other lattice-based, architected materials allows for the assignment of discrete coordinates to each possible mechanism of damaging the material.

Figure \ref{fig:placeholder2}A shows an undamaged HSA twisted to its maximum position of one full rotation. We consider in this work an HSA ``cut'' as fully severing either a vertical or horizontal auxetic link, the location of which is given by the vertical helix (A, B, C), and its height on the HSA (0-7 for vertical, 0-6 for horizontal). Burns, or melting auxetic links together, are segmented into vertical coordinates of (0-2) given the imprecise nature of melting HSAs. In total, this provides 54 different possible ways of damaging an HSA. We assume HSAs to be non-functional in their ability to generate forces and torques when more than 3 instances of damage are applied. The damage state of an HSA $x$ is then represented as a sparse binary vector: 

\begin{equation}
    x \in \{0,1\}^{54}, \qquad \|x\|_0 \le 3.
\end{equation}

While this specific structure of vertical and horizontal struts is not universal to all architected materials, this coordinate system provides a physically grounded mechanism for engineering adaptive behaviors in auxetics and other architected materials. The goal of this encoding is not to explicitly and accurately model every combination of possible HSA damages. Rather, the goal is to produce an encoded representation such that the distribution of latent variables $z$ produces ensembles of dynamically feasible force/torque predictions. Similar parameterizations are readily available for a wide range of architected materials \cite{spinoid_architected, Li2023}. Since LEAP uses ensemble methods, full coverage of the latent space, while desirable, is not necessarily required. The goal is kinematically feasible and reasonably orthogonal basis clusters, not comprehensive coverage.

\subsection{Modeling Damage via a Latent Space} \label{latent modeling}

We model the damage state of the architected material using a variational latent model, an architecture that produces smooth probabilistic latent spaces that act as a compressed representation of its input data. As described in Section \ref{damage_as_coords}, we transform the discrete coordinate representation of damaged links in the HSA into a sparse binary vector where the damaged coordinates are represented by ones. We then learn the latent space mapping (Eq. \ref{equation: d}).

\begin{equation}
    z = \mathcal{D}(x), \qquad \mathbb{R}^{54} \to \mathbb{R}^{3}
    \label{equation: d}
\end{equation}

The final layer of $\mathcal{D}$ is parametrized as a multivariate Gaussian with means $\mu$ and a diagonal variance matrix whose non-zero elements are $\sigma$. This is then sampled to produce a latent vector $z$ using the ``re-parametrization trick'' to allow proper gradient flow (Eq. \ref{equation: z}).

\begin{equation}
    z = \mu(x) + \sigma(x) \odot \epsilon, \qquad \epsilon \sim \mathcal{N}(0,I)
    \label {equation: z}
\end{equation}

The latent model is trained using a variant of the $\beta$-VAE formulation \cite{higgins2017betavae} of the Evidence Lower Bound (ELBO) objective (Eq. \ref{equation: loss function}).

\begin{equation}
    \mathcal{L}(\mathcal{W},D;y) = \mathbb{E}_{q_\mathcal{D}(z|x)}[\log p(y|z)] - \beta D_{KL}(q_{\mathcal{D}}(z|x) \parallel p(z))
    \label{equation: loss function}
\end{equation}
where $y$ is a ground-truth force/torque measurement and $P \in SE(3)$ is the homogeneous transformation measured from the base to the top of the HSA. The first term in the objective is mean-squared error loss and $D_{KL}$ represents the Kullback-Leibler Divergence (KL-D) using a $\beta$-annealing similar to \cite{higgins2017betavae}.

The wrench predictor learns the function:

\begin{equation}
    \hat{y} = \mathcal{W}(P|z), \qquad \mathbb{R}^{12} \to \mathbb{R}^{6},
\end{equation}
where the input to the wrench predictor is the latent state $z$ concatenated with a representation of the orientation component of $P$ using a Gram-Schmidt-like process according to \cite{zhou_continuity_2020}, which performs the mapping:

\begin{equation}
g_{\mathrm{GS}}\!\left(
\begin{bmatrix}
\mid &        & \mid \\
p_1  & \cdots & p_3 \\
\mid &        & \mid
\end{bmatrix}
\right)
=
\begin{bmatrix}
\mid & \mid \\
p_1  &p_{2} \\
\mid & \mid
\end{bmatrix}
\end{equation}

This operation truncates the last column in $p$, which is the $SO(3)$ component of $P$, ensuring continuity of rotation. The wrench predictor is trained according to Eq. \ref{equation: loss function}, which in practice simplifies to mean-squared error $\|y-\hat{y}\|^2$ assuming a unit variance.

\color{black}
Table \ref{hsas_used_table} describes the distribution of the damaged HSA data provided to the pretrained LEAP model. Ground-truth data is collected via an OptiTrack fiducial tracker and a Robotous RFT60-HA01 six-axis force/torque sensor, both at 360Hz, by hand using a handle affixed to the end of an HSA. Training and testing data for each HSA includes bending, twisting, compression, extension, and buckling behaviors. Collecting a diverse set of training data is important for producing a structured latent space: while the number of combinations for 2+ types of HSA damage is large, we train on data from HSAs with one cut on each of the 21 ``vertical cut" coordinates, and each of the 9 possible burn locations.

\begin{table}[h]
\centering
\captionof{table}{Data Quantity for Training}
\vspace{-5pt}
\begin{tabular}{|c|cl|cl|}
\hline
\multicolumn{1}{|l|}{}                         & \multicolumn{2}{c|}{\textbf{Data (Minutes)}} & \multicolumn{2}{c|}{\textbf{HSAs Used}} \\ \hline
\textbf{No Damage} & \multicolumn{2}{c|}{112} & \multicolumn{2}{c|}{32} \\ \hline
\textbf{One Cut}   & \multicolumn{2}{c|}{146} & \multicolumn{2}{c|}{32} \\ \hline
\textbf{Two Cuts}  & \multicolumn{2}{c|}{27}  & \multicolumn{2}{c|}{6}  \\ \hline
\textbf{One Burn}  & \multicolumn{2}{c|}{41}  & \multicolumn{2}{c|}{9}  \\ \hline
\multicolumn{1}{|l|}{\textbf{2+ Cuts / Burns}} & \multicolumn{2}{c|}{45}                      & \multicolumn{2}{c|}{10}                 \\ \hline
\end{tabular}
\label{hsas_used_table}
\end{table}

\subsection{Adapting to Damage with HSA Ensembles}

\begin{algorithm}
    
\caption{Latent Ensemble Adaptation}
\SetKwInput{Input}{Input}
\SetKwInput{Output}{Output}
\Input{Networks $\mathcal{D}$, $\mathcal{W}$, Poses $P$, Ref. Wrenches $y$}
\Output{Ensemble Weights $\mathcal{A}$, Wrenches $\mathcal{F}$}
\DontPrintSemicolon
\BlankLine
% Initialize Latent Ensemble $E$ \;
% Initialize Ensemble Weights $W \in \mathbb{R}^{E \times m}$ \;
% Generate $e$ binary vectors $x^{(e)} \in \{0,1\}^{54}$ \; 
\text{Generate $n$ binary vectors and latent coordinates:} \;
$x_i \sim \mathcal{U}(\{0, 1\}^{54}), \quad i = 1, \dots, n$ \;

$\hat{z}_{(i)} \gets \mathcal{D}(x_i)$ \; %\text{Generate latent states} \;

\BlankLine

\text{Construct latent ensemble via clustering:} \;

% Obtain ensemble of $E$ latent states using the centroids of k-means clustering $\{z_1, \dots, z_E\}$ \;
$E \leftarrow \text{K-Means}(\{\hat{z}_1, \dots, \hat{z}_n\}) \text{ where } E = \{z_1, \dots, z_K\}$\;

\BlankLine 
\text{Generate ensemble predictions:} \;

\For{$i = 1$ to $K$}{
    $\hat{y}_{(i)} \gets \mathcal{W}(P|z_i)$}

$\hat{Y} = [\hat{y}_{(1)}, \dots, \hat{y}_{(E)}]$ \;

\BlankLine 

\text{Solve least-squares regression:} \;

$\mathcal{A} \gets \arg\min_{\mathcal{A}} \| \hat{Y} \mathcal{A} - y \|_2^2$ \;
\Return{$\mathcal{A}$}

\BlankLine
\textbf{LEAP Inference:}

$\mathcal{F} \gets \mathcal{A}^T\cdot\mathcal{W}(P,E)$ \;
\Return{$\mathcal{F}$}
\end{algorithm}

\color{black}
Rapid adaptation is critical not only due to constraints on compute and time, but also on hardware---a partially-broken system is more likely to degrade during the adaptation process itself if it is sample-inefficient. LEAP overcomes this constraint by using a simple yet effective ensemble method, requiring very little data and consisting only of linear operators that regress in near real-time. Furthermore, practically functional adaptation methods must be robust to large variations in damage, rather than defining adaptable states into rigid bounds that can be easily predicted by first principles. As such, our method for adapting in-situ involves no training or fine-tuning of either the damage encoder $\mathcal{D}$ or force predictor $\mathcal{W}$.

To adapt to out-of-distribution damage, our method relies on the weighted average of an ensemble of predictions driven by the latent structure. The ensemble is produced by clustering a large number of sampled encoder outputs to produce a computationally tractable ensemble of $E$ latent coordinates $E=\{{z}_1 \dots z_K\}$ at the centroids of each k-means cluster. This approximates sampling from maximally differing HSA damages. These centroids are then used to produce an ensemble of force/torque predictions by evaluating the entire adaptation window through the wrench predictor $\mathcal{W}(P,z)$.

Given either a ground truth force/torque signal or a reference weight picked up by a manipulator, weightings by-component for each member in the ensemble are calculated using a least-squares regression. Inference after obtaining ensemble weightings is given by: 

\begin{equation}
    y(P) = \sum^{i = 30}_{i = 1}a_i \mathcal{W}(p_,z_i)
    \label{equ: ensamble}
\end{equation}

As all other elements of the ensemble are generated before runtime, regression of the ensemble weights is computationally lightweight and scales linearly with the number of actuators in the overall system. In practice, we find that for single HSAs, strong performance is achieved after 30 seconds of data collection at 100Hz, 5 seconds of data pre-processing and 2 seconds of regression, totaling 37 seconds for the whole process.

Our method does assume that the transform between HSA ends is known at runtime. For our evaluations on a 6-DoF soft wrist, this is achieved by a camera and AprilTag inside the mechanism as shown in Figure \ref{fig: exploded view}. For a single HSA, this ground truth is provided by OptiTrack motion capture cameras.
\color{black}

\section{Extending LEAP to a 6-DoF Soft Wrist}
\label{sec:wrist_adaptation_method}

As a linear ensemble method, LEAP naturally extends to arbitrary hybrid soft robots provided that:

\begin{itemize}
    \item Actuators originate and terminate exclusively between rigid elements,
    \item the relative positions of the rigid elements are known, and
    \item actuator models can be developed as described in Section \ref{latent modeling}.
\end{itemize}

If these conditions are met, the learned adaptation problem can be formulated as a linear least-squares regression over ensemble weightings---the only difference to the single-HSA case is the dimension of the linear regression. 

\begin{figure}
    \centering
    \includegraphics[width=\linewidth]{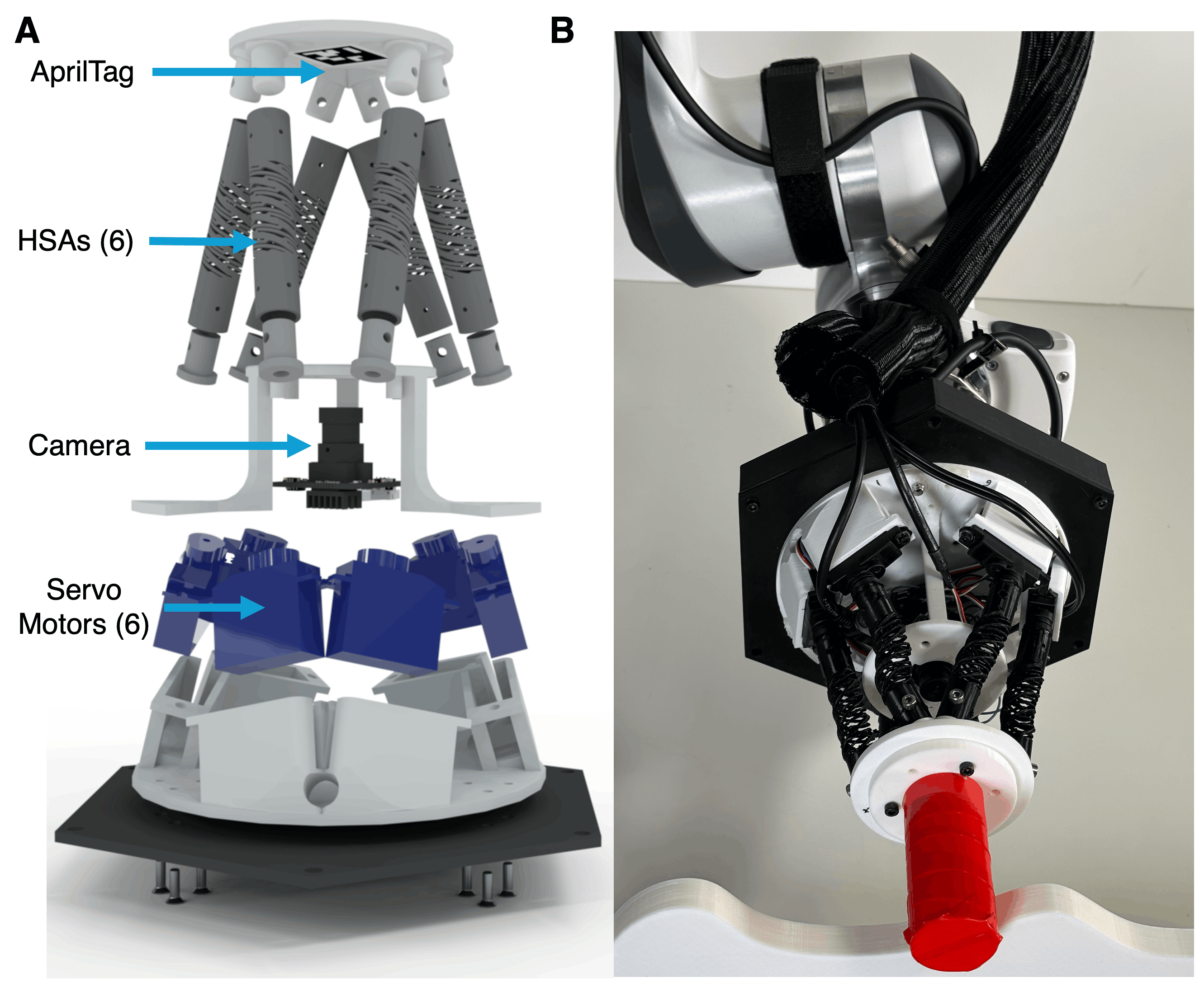}
    \caption{\textbf{HSA-Actuated Soft Wrist: }A. We demonstrate LEAP on a miniature, parallel 6-DoF soft wrist. Servomotors drive HSAs, and a proprioceptive camera determines platform pose via an AprilTag. 
    B. The soft wrist mounted to a Franka Emika Panda robot arm. We demonstrate the wrist's functionality via a tracing task where the objective is to traverse the contour of a 3D-printed surface while remaining in contact as much as possible.
    }
    \label{fig: exploded view}
\end{figure}

To evaluate our method, we designed a 6-DoF soft wrist driven by six HSAs, as shown in Figure \ref{fig: exploded view}. Each HSA is driven by a servomotor that has 180 degrees of motion, resulting in a maximum linear travel distance of 7.2mm in $Z$. Contained within the wrist is a camera and AprilTag for proprioceptive pose sensing that is used for force/torque prediction during runtime.

We begin by establishing the kinematic relationships between each HSA actuator and the two rigid components of the system: the wrist chassis and the interface plate where the AprilTag is mounted. In this work those relationships are recorded in an XACRO file based on the system's CAD model, but in principle could be determined experimentally or even learned. The internal camera establishes the position of the interface plate relative to the wrist chassis and the forward kinematics determines the transforms of each HSA relative to its base. 

This transform is then augmented with servo rotation feedback to provide the inputs for LEAP. While it is feasible to model each actuator as its own ensemble with a separate set of damage seeds, we chose instead to initialize a single set of 30 latent clusters and determine different weightings for every constituent actuator on the wrist. The wrenches generated by each element of the ensemble can then be transformed to a common frame, weighted, and summed. The adaptation process, as with a single actuator, is equivalent to performing a least squares regression over those weights.

\begin{figure*}
    \centering
    \includegraphics[width=1\linewidth]{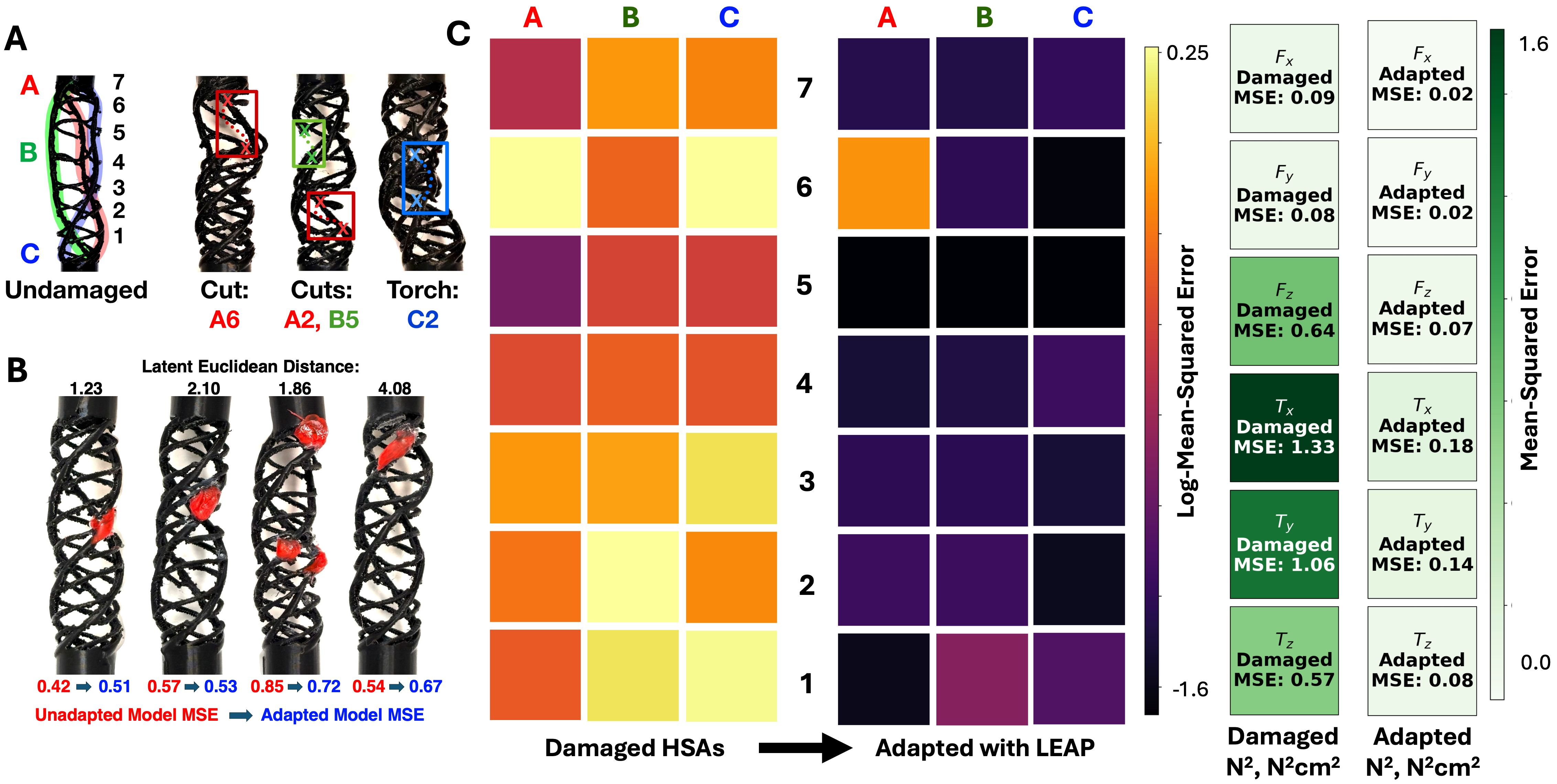}
    \vspace{-10pt}
    \caption{\textit{\textbf{LEAP Enables Robust, Rapid Damage Adaptation.}} \textbf{A.) HSA Damage as Discrete Coordinates:} The periodic structure of architected materials introduces degeneracies: in an HSA, severed links can be either on a vertical strut (A, B, C) highlighted in the undamaged HSA, or on the horizontal links connecting the major struts. Torching HSAs occurs in one of three locations along each strut. \textbf{B.) Repair in Architected Materials:} Adaptation after repair in an architected material is operationally equivalent to damage adaptation and achievable by hobbyists with limited resources. Cut HSAs are repaired with glue and after adaptation their weighted latent coordinates are distinct despite all links being intact. \textbf{C.) Adaptation across the space of possible cuts.} We perform a vertical cut at every coordinate along an HSA. After evaluating LEAP, we observe a performance improvement of 87\% in MSE loss across all components. Average component error decreases by 85\% after adaptation.}
    \label{fig:placeholder2}
\end{figure*}

\section{Results}

Our experiments were structured to answer the following questions:
\begin{itemize}
    \item How well does LEAP adapt to damage in a single HSA?
    \item Does LEAP extend to systems with multiple HSAs in parallel, such as a 6-DoF wrist?
    \item How well does LEAP generalize to out-of-distribution changes, including unseen damages and actuator repairs?
    \item Can LEAP adaptation produce task-level improvements?
\end{itemize}
\color{black}

\begin{figure}
    \centering
    \includegraphics[width=\linewidth]{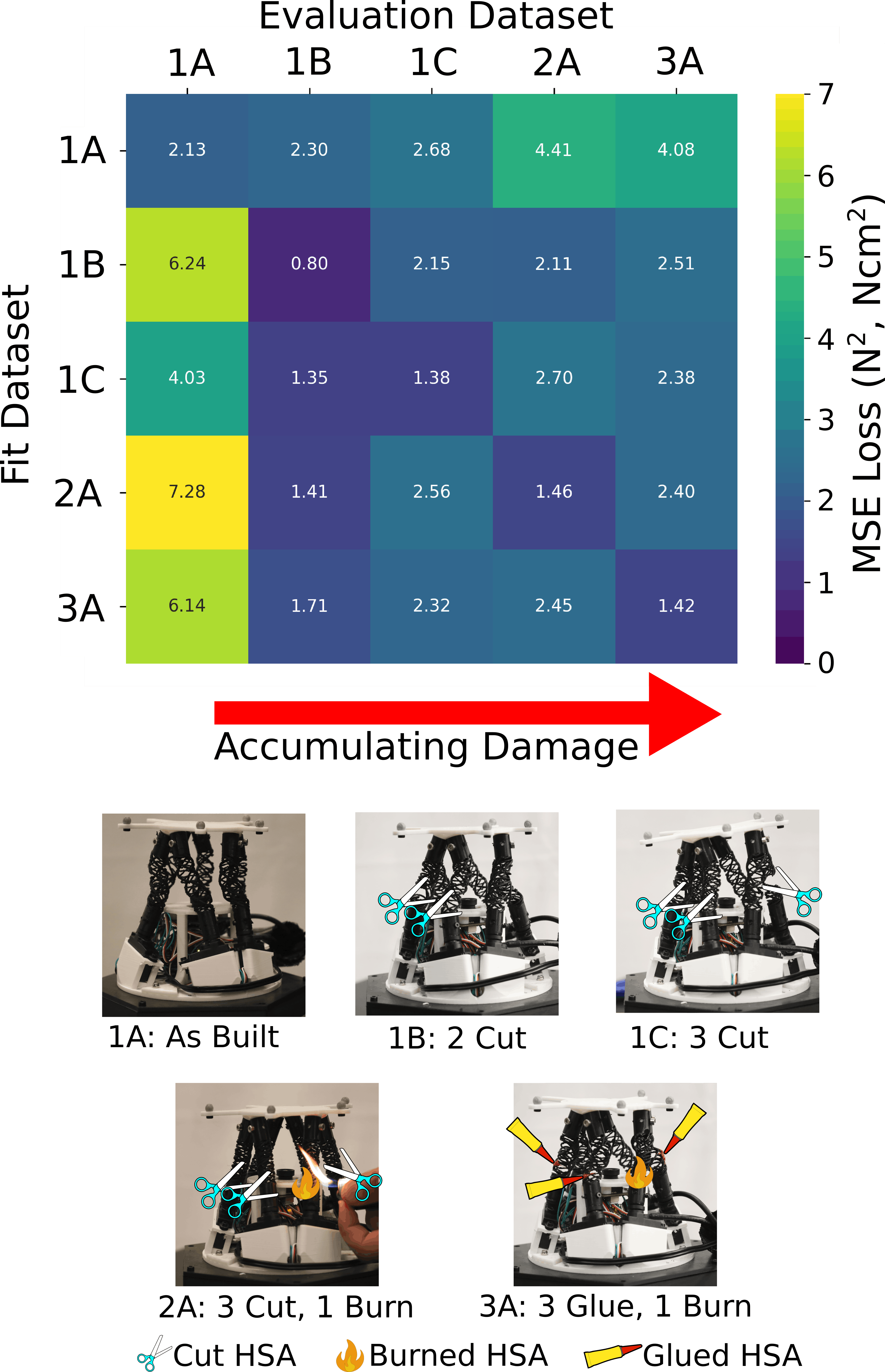}
    \caption{\textbf{Wrist Adaptation Performance: } Our whole system model fit to data from increasingly damaged wrist configurations. The confusion matrix diagonal shows that the system is able to adapt to a variety of system damage states using less than a minute of total data collection and computation time. The off-diagonal elements show a clear trend whereby increasing damage accumulation degrades model performance in the absence of adaptation.}
    \label{fig: wrist adaptation}
    \vspace{-10pt}
\end{figure}

\subsection{Single HSA Damage Adaptation}
In Figure \ref{fig:placeholder2}, we show the impact of a vertical cut at every coordinate across an HSA. Figure \ref{fig:placeholder2}A indicates the coordinate system used to describe where cuts and burns along an HSA are located. We validate LEAP by comparing a damage-adapted HSA to itself with LEAP weights learned for the fully-intact actuator (i.e., after damage occurs, but before adaptation). Each square in Figure \ref{fig:placeholder2}C represents a cut at the corresponding coordinate in an HSA---we compare each cut to a prior evaluation of each HSA to control for wear and manufacturing inconsistencies.

Without adaptation, damaged HSAs have force/torque sensing errors of as much as 0.25 (N$^2$, N$^2$cm$^2$) log-MSE, or 1.78 (N$^2$, N$^2$cm$^2$) mean-squared error averaged per component. After adaptation, LEAP produces an average reduction in sensing error of 87\% compared to the damaged and unadapted HSAs.

While LEAP improves performance across all axes, the largest raw improvements occur in $T_x$, $T_y$, $F_z$, and $T_z$. This is largely due to the geometry of the architected transmission: HSAs transduce torques and forces along the $Z$ axis, and the beam bending behavior of HSAs along with the reaction forces due to the material's handedness produces torques about $X$ and $Y$. As shown in Supplemental Figure S1, unadapted predictions can be accurate in the $X$ and $Y$ axes but highly inaccurate in $Z$, the primary force/torque transmission axis.

\begin{figure}
    \centering
    \includegraphics[width=\linewidth]{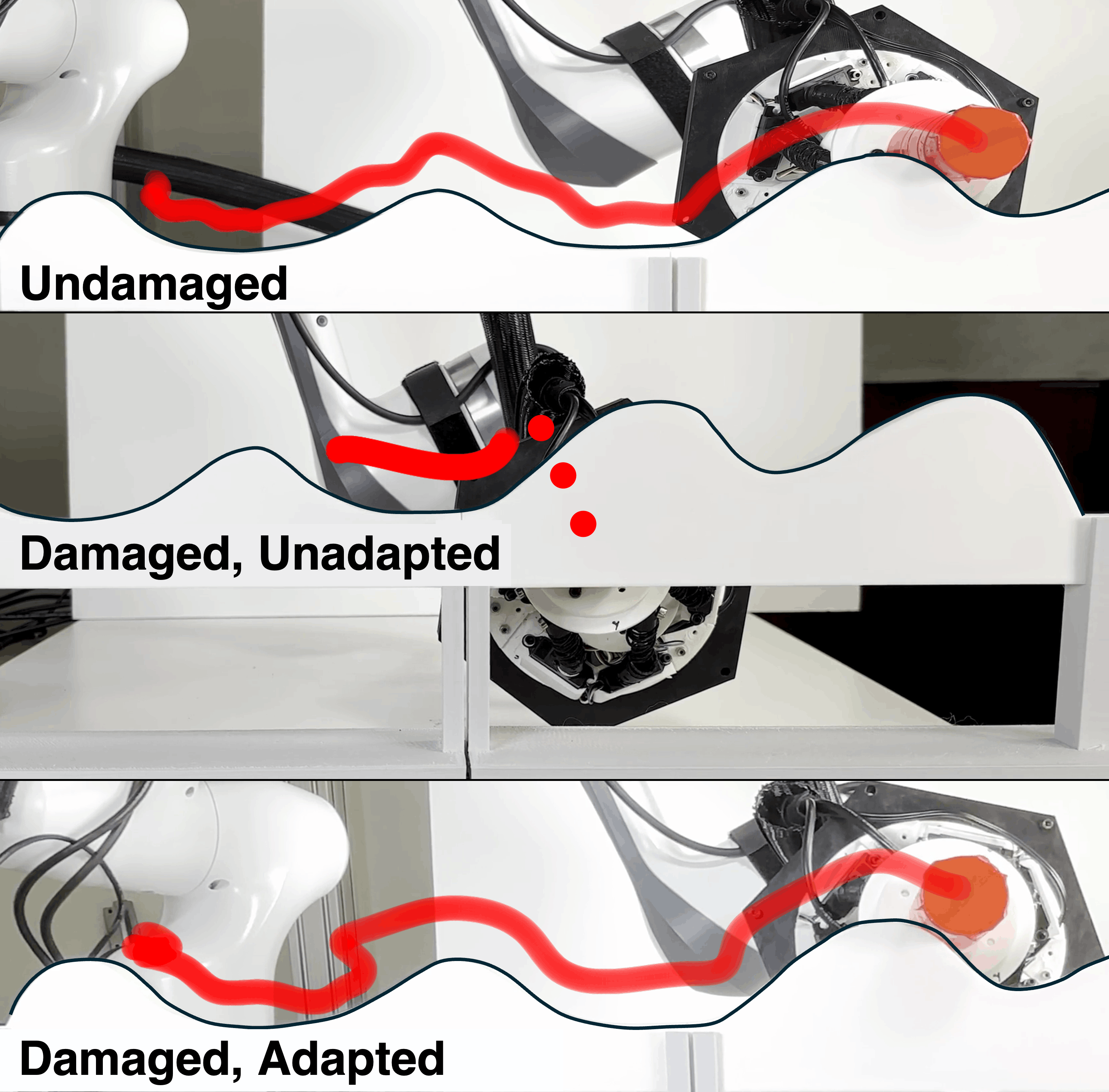}
    \caption{\textbf{Tracing Task Demonstration: } Representative trajectories from the proprioceptive tracing task. Without adaptation, the damaged wrist is not task capable, it fails to follow the contour and cannot recover. After adapting with LEAP, the system recovers up to 100\% of its original performance.}
    \label{fig: tracing eval}
    \vspace{-10pt}
\end{figure}

\subsection{Adaptation on the 6-DoF Soft Wrist}

We used LEAP to adapt force/torque proprioception models to five different configurations of our HSA-actuated soft wrist system, as shown in Figure \ref{fig: wrist adaptation}. Beginning with an undamaged robot, we incrementally introduce cuts, burns, and repairs to modify the dynamics of the system. In each case, we adapt a LEAP proprioception model to 60s of randomly selected data, a process which took 4.5s on an RTX3080 laptop GPU. The ground-truth for this experiment was provided by the six-axis force-torque sensor and motion tracking system described in Section \ref{sec:wrist_adaptation_method}. 

We then evaluate each resulting model on data from all five robot configurations and present the results in Figure \ref{fig: wrist adaptation}. Along the diagonal of the confusion matrix in Figure \ref{fig: wrist adaptation}, the LEAP ensemble weights are fit to the specific damaged embodiment. In the top-right, LEAP is evaluated on embodiments that are more damaged than its weights were fit to, in the bottom-left the LEAP weights are fit to embodiments that are more damaged than what is evaluated. 

We find that LEAP is able to model all five robot configurations, with mean error magnitudes of 1.5N in force and 0.074Nm in torque. LEAP delivers significant improvements---adapted models decrease error by an average of 49\% compared to models trained on another robot configuration and deployed without adaptation. Figure \ref{fig: wrist adaptation} shows that unadapted models generally perform better in damage configurations that produce force/torque distributions that are closer to the support of their original LEAP adaptation data. However, it is interesting to note that the biggest drop in performance came from deploying models trained on the fully undamaged system on partially damaged datasets. We believe this reflects the additional stiffness imparted by having six undamaged actuators in a near-kinematically-constrained arrangement.

\subsection{Adaptation to Repaired HSAs}

Repair of architected materials such as HSAs introduces inconsistencies that must be accommodated with an adaptation procedure such as LEAP.

In Table \ref{tab:hsa springs}, we measure the Z-axis spring constant of five HSAs---as-printed, after a vertical cut, and after the auxetic links were repaired using glue. Arrows indicate whether the repair increased or decreased the spring constant compared to the intact HSA. Repair is an imprecise process: despite repairing each HSA, their spring constants differed from the default stiffness by 9\% in varying directions.

\begin{table}[h]
\centering
\captionof{table}{HSA Spring Constants Before And After Repair}
\vspace{-5pt}
\label{tab:hsa springs}
\begin{tabular}{|c|r|r|r|l|}
\hline
\multicolumn{1}{|l|}{} & \multicolumn{1}{c|}{\textbf{Default}} & \multicolumn{1}{l|}{\textbf{One Cut}} & \multicolumn{1}{l|}{\textbf{Repaired}} & \textbf{Unit} \\ \hline
\textbf{HSA 1}                       & 0.36 & 0.30  & $\uparrow$ 0.39 & N/mm \\ \hline
\textbf{HSA 2}                       & 0.33 & 0.30  & $\uparrow$ 0.36 & N/mm \\ \hline
\textbf{HSA 3}                       & 0.34 & 0.30  & $\downarrow$ 0.32 & N/mm \\ \hline
\textbf{HSA 4}                       & 0.33 & 0.27 & $\uparrow$ 0.35 & N/mm \\ \hline
\multicolumn{1}{|l|}{\textbf{HSA 5}} & 0.34 & 0.29 & $\downarrow$ 0.30  & N/mm \\ \hline
\end{tabular}
\end{table}

This behavior extends to LEAP's latent representations---in Figure \ref{fig:placeholder2}B we compare four undamaged HSAs to their repaired counterparts. While LEAP performs similarly in mean-squared error for force/torque prediction compared to the unadapted case, the weighted latent coordinates for the healthy and damaged ensembles differ significantly, indicating that a change occurs in the HSA dynamics. We explicitly made the choice to not include repaired HSAs as a means of testing out-of-distribution performance---in practical applications, actuator repair is an ad-hoc process in response to unpredictable damage, which occurs in the context of the environment and resources available.

\subsection{Task Demonstration}

To demonstrate LEAP's performance in a contact-rich setting, we designed a tracing task wherein the HSA-actuated soft wrist is mounted to a Franka Emika Panda robot arm. The objective is to use the proprioception of the wrist alone (i.e., no vision or predefined motions) to slide along a curved contour, maintaining contact as much as possible.

In this setting, the wrist is passive and functions only as a sensor. For tracing, it is only required that the proprioceptive outputs are ``force-like" and ``torque-like" (i.e., have a monotonic relationship with respect to the real forces and torques experienced by the system). However, they must be directionally correct and maintain a consistent magnitude after adapting to damage, as the motion of the arm is driven by a PID loop about the wrist's sensory output.

Before the wrist is damaged, the system is able to keep the end of the wrist in sliding contact with the contour for 85\% of the total trajectory. After adapting to burning one HSA and cutting a second (see the provided multimedia), the wrist is able to stay in contact with the contour for an identical 85\% of the total path averaged over three trials, measured using a video analysis tool. Without LEAP adaptation, the damaged wrist maintains contact only briefly and catastrophically fails before reaching the end of the contour path---the damaged, unadapted system is not task capable.

\color{black}

\section{In-Distribution Properties}

\label{damage linearity}

One of the insights from this work is that architected materials have a number of advantages over conventional continuum soft materials when it comes to characterizing and adapting to damage. 

For truss-based architected materials, such as HSAs, cuts at different points along a discrete element of the structure are functionally degenerate, creating the same impact on the bulk material behavior. This observation forms the basis of first-principles truss analysis, used to characterize the behavior of infrastructure such as bridges, buildings, and towers. For linear elastic structures, the principle of superposition dictates that deflections caused by multiple forces can be summed independently of their ordering \cite{linearelasticity}---this implies that a finite number of destructive tests would cover the distribution of all damages.

In an architected material, these same degeneracies create a coordinate system by which damage can be encoded and damages can be superimposed. For example, in the HSAs used in this paper, any cut, burn, or combination thereof can be accurately described as part of a 54 dimensional binary vector.

We considered three types of damages: Removing a link in the structure (cutting), restoring a link in the structure (gluing), and fusing adjacent links (burning): all of these either increase or decrease the stiffness of the architected material. Since these changes are local, the damage accumulated in one part of the structure does not change the impact of additional damage elsewhere in the structure. Moreover, the impact of multiple damages is therefore order-independent, as long as they are non-overlapping \cite{linearelasticity}.

Therefore, multiple instances of damage can be understood as a linear combination of their individual effects. This property is important for sample efficiency \cite{kohn2001nonparametric, korda2020optimal} and, in effect, LEAP learns the impacts of individual damages along with residuals of their linear combinations. The approximate linearity of damage states on an HSA also motivates the choice of ensembles of wrench predictions. We rely on the support of the ensemble distribution to encompass the ground-truth signal and expect linear combinations of HSA damages to be supported under a data collection process where largely only single HSA damages are considered.

An architected material with a discrete damage description is intrinsically useful: a finite number of experiments can fully model the excitation of any damaged mechanism with respect to its learning inputs. For discrete damage, fully modeling the system requires $N^m$ datasets, where $N$ is the size of the binary damage vector and $m$ is the number of times each transmission can be damaged.

However, superimposing linear combinations of damage collapses this exponential scaling. In a perfect linear elastic truss, only $N$ datasets are required to fully model the system. In practice, we find that additional data is helpful for model stability---particularly in modeling the process variability of HSA manufacturing---though we consider $N\times m$ to be an approximate upper bound for the number of experiments required. Additionally, when designing HSA-actuated mechanisms, the quantity of HSA active units changes with either length or width (diameter) of the transmission, but not both. This means that the scaling complexity of $N$, and therefore the binomial distribution of possible damages, is linear with respect to changes in design.

This has implications for robots that expect to sustain and adapt to damage. In contrast to rigid or continuum soft mechanisms, in-distribution characteristics are guaranteed for learning representations of damaged architected materials given a finite and countable number of experiments. The ability to employ LEAP and LEAP-like algorithms reflects an inherent strength of architected materials relative to other classes of actuator.

\section{Conclusion}

In this work we introduce LEAP, a method of adapting to unseen damages in seconds in architected materials. We identify properties of architected materials that can be approximated with a truss-like structure that provides in-distribution guarantees for all local damages while using a comparatively small quantity of data and a countable number of physical experiments---a distinct advantage over both rigid mechanisms and continuum soft actuators. We evaluate LEAP in single- and multi-HSA systems, introducing a 6-DoF soft wrist, and demonstrate that LEAP recovers proprioception performance in a tactile tracing task.

An additional empirical finding from our experiments in single- and multi-HSA systems, as shown in Figure \ref{fig: wrist adaptation} and supplementary Figure S1, is that improved proprioceptive results can be achieved when portions of the architected materials were broken. Given the nonlinearities present in HSAs, separating the energy-containing members of its auxetic structure provides an explanation for why this is: by removing nonlinearities, the architected material bears a closer resemblance to a linear spring or simple beam, which can be more easily parameterized. Architected materials possess an advantage in adaptability not only from their in-distribution properties as discussed in Section \ref{damage linearity}, but also from the selective removal of structure in order to improve a behavior or assume a new one altogether. Further investigation of the design tradeoff between force transmission and algorithmic observability is a direction of future work.

\color{black}

Limitations do exist to our method: one drawback is that despite approximately linear experimental complexity, a reasonable amount of actuator characterization is required in order to apply LEAP. Furthermore, there exists room for improvement in the ensemble selection process itself---our method presumes that selecting centroids of the latent clusters provides sufficient dynamical variation to solve the least-squares weighting. More intelligent selection of ensemble members or specific motion strategies for collecting adaptation data would both improve efficiency and performance. Another limitation is that we do not encode damages such as wear, abrasion, and localized hinging in this work, although anecdotally we find that LEAP is able to adapt well to all three, as well as HSA manufacturing variability.

This paper presents not only a method of adapting to damage and degradation of robotic components but also a new lens through which to view robot design. Representing the damage state of an architected material as a discrete encoding satisfies conditions that are useful for statistical learning beyond proprioception and self-modeling. \textbf{One interpretation of this work is that architected materials---because of their in-distribution guarantees---have advantages over rigid and continuum materials for mechanisms that will be required to learn.}

\section{Acknowledgments}

We thank Ayush Gaggar for his insightful feedback on this work. This research is based upon work supported by the US Army Research Office Grant no. W911NF-22-1-0286 and the National Science Foundation under Grant no. 2330040, the Engineering Research Center (ERC) for Human Augmentation via Dexterity (HAND). J.A. is supported by a National Defense Science and Engineering Graduate Fellowship.

\bibliographystyle{unsrtnat}
\bibliography{references2}
\balance
% \newpage

\begin{appendix}
\section{sec: appendix}

\subsection{Case Study of Twice-Damaged HSAs}

\begin{figure*}[hbtp] 
    \centering
    \includegraphics[width=0.7\linewidth]{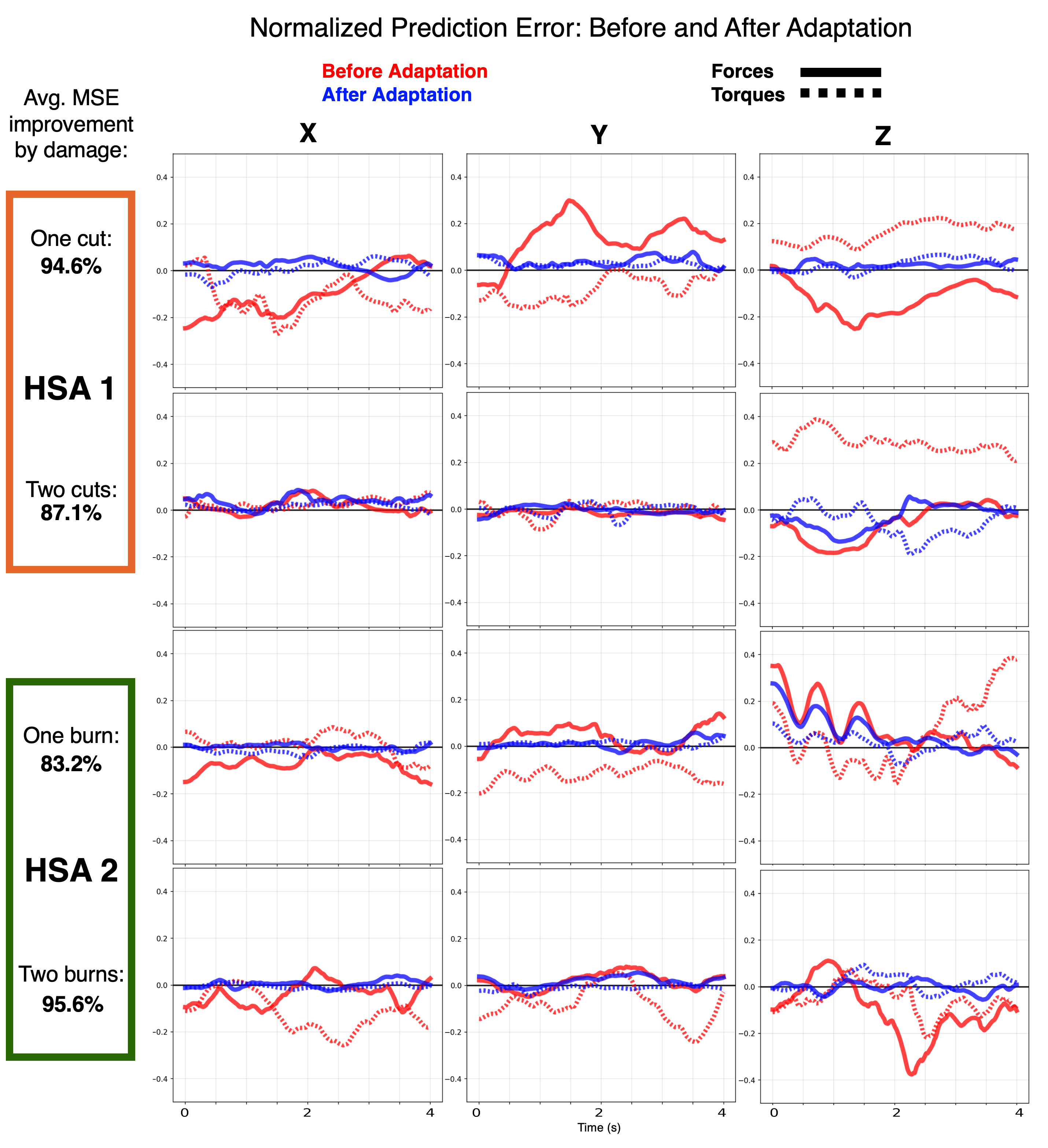}
    \caption{\textbf{Normalized Error Before and After Adaptation:} Pictured is a representative sample of force-torque errors over time---closer to zero is better. Data is plotted for two HSAs that each undergo multiple damages sequentially. }
    \label{fig: error plot}
\end{figure*}

Figure \ref{fig: error plot} shows force-torque prediction errors over time for two HSAs that each undergo two damages sequentially; improved performance is closer to zero. As HSAs produce forces in response to torques about the Z-axis, the normalized error is higher in those dimensions of the predicted wrenches. Furthermore, as cuts remove nonlinearities from the HSA structure, prediction accuracy can in fact improve as the material is damaged.

\subsection{Transmissions as Distribution Mappings}

Architected materials provide a transmission of forces and torques. However, compared to fully-rigid structures, architected transmissions 
provide a mechanism for transforming a distribution of actuator commands $\mathcal{X}$ to a distribution of wrenches $\mathcal{F}$ (on a robot, for example); how a distribution is transformed varies dramatically with the metamaterial architecture. This property is important in statistical learning, particularly in embodied learning, where the performance of a physical learner is coupled to the distribution of inputs available to it during the learning process. For example, dimensionality of the excitation distribution, its similarity to the state distribution of the system during operation, and how close the resulting data set is to being independent and identically distributed (\textit{i.i.d.}) \cite{wu2022daydreamerworldmodelsphysical, Berrueta2024} all impact learning outcomes. \ As a result, changes to 
an architected material may create output distributions better suited to learning \cite{pfeifer2006body}.  

\subsection{Soft Wrist Technical Specifications}

The engineered system described in this work is a 6-DoF soft wrist designed to provide a mechanical adaptation layer for learning systems engaged in dexterous tasks. It features a ``lower" platform that connects to a robot arm, and an ``upper" platform with a bolt-hole pattern for end effectors. Kinematically, the soft wrist approximates a compliant Stewart-Gough platform, with six HSA transmissions providing both actuation and structure to the upper platform.

The platform has a total height of 115mm, a base diameter of 130mm, and a total weight of 400g.  For experimental convenience we chose to use only clockwise HSAs in this work, which somewhat reduced the platform's range of motion and introduced additional deviation from a rigid dynamics approximation relative to an alternating handedness configuration. The platforms range of motion as-configured can be found in Table \ref{tbl: range of motion}. The platform has a full-length stroke time of 0.2s.

\begin{table}[h]
\centering
\captionof{table}{Platform Range of Motion}
\vspace{-5pt}
\begin{tabular}{|c|cl|cl|}
\hline
\multicolumn{1}{|c|}{Axis} & \multicolumn{2}{c|}{\textbf{Stroke}} & \multicolumn{2}{c|}{\textbf{Unit}} \\ \hline
\textbf{X} & \multicolumn{2}{c|}{15.5} & \multicolumn{2}{c|}{mm} \\ \hline
\textbf{Y}   & \multicolumn{2}{c|}{15} & \multicolumn{2}{c|}{mm} \\ \hline
\textbf{Z}  & \multicolumn{2}{c|}{7.5}  & \multicolumn{2}{c|}{mm}  \\ \hline
\textbf{Roll}  & \multicolumn{2}{c|}{0.08}  & \multicolumn{2}{c|}{Rad}  \\ \hline
\textbf{Pitch}  & \multicolumn{2}{c|}{0.1}  & \multicolumn{2}{c|}{Rad}  \\ \hline
\textbf{Yaw}  & \multicolumn{2}{c|}{0.2}  & \multicolumn{2}{c|}{Rad}  \\ \hline
\end{tabular}
\label{tbl: range of motion}
\end{table}

\subsubsection{Kinematics}

\begin{figure*}[ht]
        \centering
        \includegraphics[width=.9\linewidth]{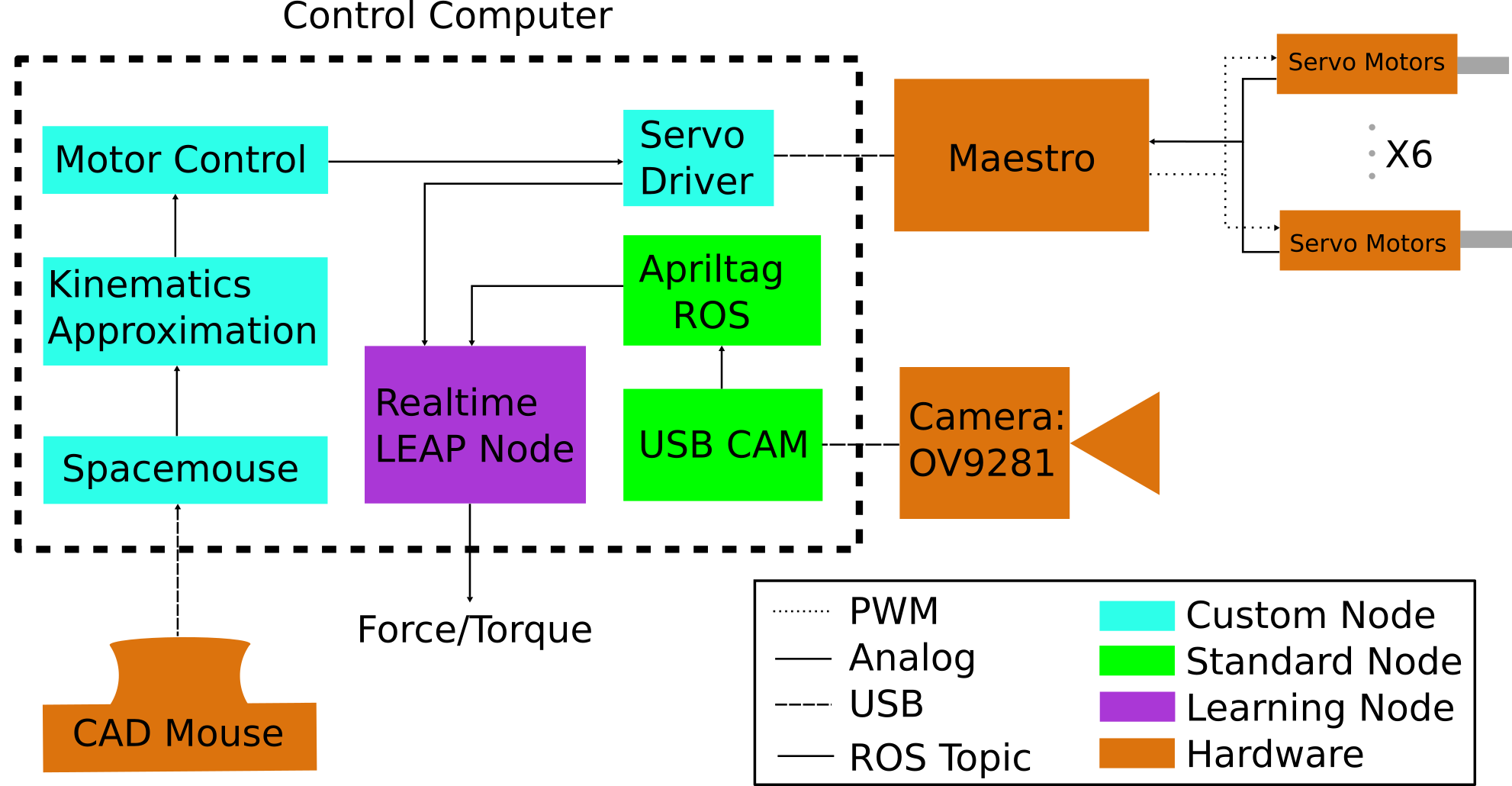}
        \caption{\textbf{Wrist Block Diagram}: The wrist is controlled and interfaced to LEAP via ROS2, associated nodes and hardware are shown.}
        \label{fig:system diagram}
\end{figure*}

A conventional rigid Stewart platform can be kinematically constrained with four values: $a_l$ the angle from each anchor point to its associated corner on the lower platform, $a_u$ the angle from each anchor point to its associated corner on the upper platform, $r_l$ the lower anchor point radius and $r_u$ the upper anchor point radius. However, since this is a hybrid soft system, we have provided several additional values which would be required for replication: $l_{hsa}$, the length of each HSA at rest, $t_l$/$t_u$ the tilt of each anchor point along its associated side, and $c_l$/$c_u$ the camber of each anchor point perpendicular to its associated side.  These values can be found in Table \ref{tbl: kinematics}.

\begin{table}[h]
\centering
\captionof{table}{Platform Kinematics Parameters}
\vspace{-5pt}
\begin{tabular}{|c|cl|cl|}
\hline
\multicolumn{1}{|c|}{Axis} & \multicolumn{2}{c|}{\textbf{Stroke}} & \multicolumn{2}{c|}{\textbf{Unit}} \\ \hline
\textbf{$r_l$}  & \multicolumn{2}{c|}{45}  & \multicolumn{2}{c|}{mm}  \\ \hline
\textbf{$r_u$}  & \multicolumn{2}{c|}{29}  & \multicolumn{2}{c|}{mm}  \\ \hline
\textbf{$l_{hsa}$}  & \multicolumn{2}{c|}{68}  & \multicolumn{2}{c|}{mm}  \\ \hline
\textbf{$a_l$} & \multicolumn{2}{c|}{0.7} & \multicolumn{2}{c|}{Rad} \\ \hline
\textbf{$a_u$}   & \multicolumn{2}{c|}{0.15} & \multicolumn{2}{c|}{Rad} \\ \hline
\textbf{$t_l$}  & \multicolumn{2}{c|}{0.26}  & \multicolumn{2}{c|}{Rad}  \\ \hline
\textbf{$t_u$}  & \multicolumn{2}{c|}{0.26}  & \multicolumn{2}{c|}{Rad}  \\ \hline
\textbf{$c_l$}  & \multicolumn{2}{c|}{-0.26}  & \multicolumn{2}{c|}{Rad}  \\ \hline
\textbf{$c_u$}  & \multicolumn{2}{c|}{0.26}  & \multicolumn{2}{c|}{Rad}  \\ \hline
\end{tabular}
\label{tbl: kinematics}
\end{table}

\subsubsection{Control and Electronics}

The soft wrist is actuated by six Feetech FT1117M-FB servos that are driven by a Mini Maestro 18 servo driver, which also measures position feedback from each of the servos. An 0V981 UVC Arducam with an M12 lens provides Apriltag tracking at 100hz and 720p. Video data is ingested using the usb\_cam node, and processed using the apriltag\_ros node which publishes a real time lower-to-upper platform transform to the tf tree. The apriltag is tag25h9:0 at size 0.0078m and is adhered to the underside of the upper platform. 

During manual testing the system is controlled with a 6DOF SpaceMouse Compact from 3DConnexion. The resulting poses are then converted to motor commands by a rigid approximation of the inverse kinematics. Previous work on the control of soft Stewart platforms suggests that this kind of rigid approximation can be sufficient for course-grain control of similar mechanisms \cite{11020875}. We found this to be true for the unloaded case. However, we did observe larger deviations when the platform was loaded, in part due to the choice to use the same, as opposed to alternating handedness for the HSAs. The control and computation for the wrist were provided by an external laptop running ROS2. Figure \ref{fig:system diagram} shows a block diagram of the control for the engineered system.

\begin{figure*}[htbp]
        \centering
        \includegraphics[width=.8\linewidth]{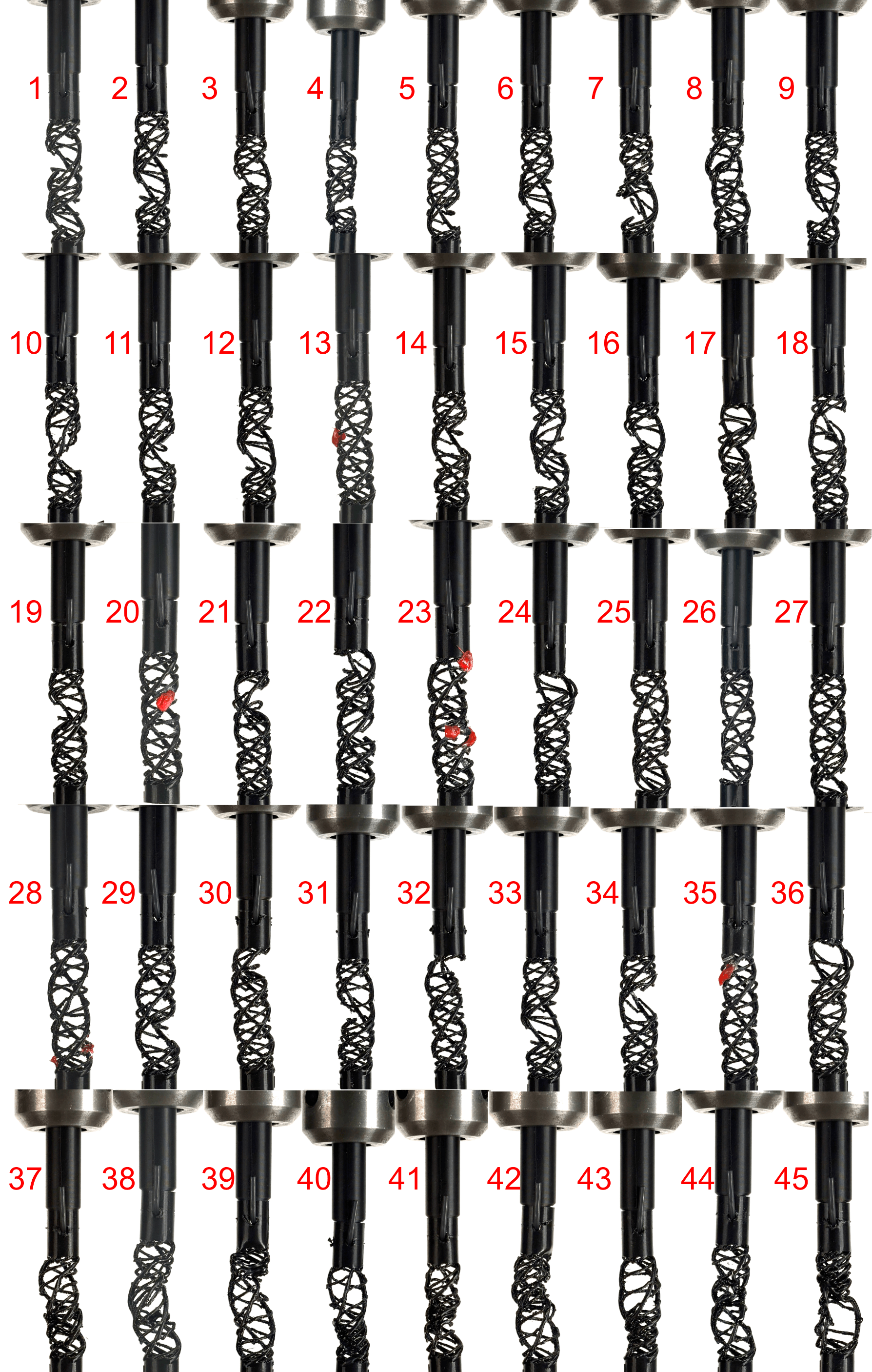}
        \caption{\textbf{All HSAs used in this work.} Table \ref{tbl: all hsas} shows how each HSA was damaged.}
        \label{fig:all hsas}
\end{figure*}

\subsection{HSA Design}

The Handed Shearing Auxetic actuators have an inner diameter of 10mm and an outer diameter of 12.5mm, and vertical struts A, B, and C are additionally reinforced to a total diameter of 14.5mm. All of the HSAs used in this work are of counter-clockwise handedness and were printed out of FormFutura Centaur polypropylene on a Prusa XL without supports. After printing, the HSAs were post-processed with a hobby knife to free any fused links.

\begin{table*}[t]
\centering
\caption{\textbf{All HSAs used in this work, and each of their damage conditions.} `VC' indicates a vertical cut, `HC' indicates a horizontal cut, and `B' indicates a burn. Highlighted rows underwent repairs.}
\begin{minipage}{0.48\textwidth}
\centering
\begin{tabular}{|c|c|c|c|}
\hline
\textbf{HSA} & \textbf{Damage 1} & \textbf{Damage 2} & \textbf{Damage 3} \\
\hline
1 & VC A1 & HC B3 & VC B4 \\
2 & VC C3 & - & - \\
3 & VC A4 & - & - \\
4 & B C0 & VC B3 & VC A7 \\
5 & VC C1 & - & - \\
6 & VC B2 & - & - \\
7 & VC B1 & VC C5 & B A1 \\
8 & B A2 & - & - \\
9 & VC C2 & - & - \\
10 & VC A3 & - & - \\
11 & VC B3 & - & - \\
12 & VC C3 & - & - \\
\rowcolor{gray!20} 13 & VC B4 & - & - \\
14 & VC B4 & - & - \\
15 & VC A5 & - & - \\
16 & VC B5 & - & - \\
17 & VC C5 & - & - \\
18 & VC A6 & - & - \\
19 & VC B6 & - & - \\
\rowcolor{gray!20} 20 & VC B6 & - & - \\
21 & VC C6 & - & - \\
22 & VC A7 & VC A0 & VC A4 \\
\rowcolor{gray!20} 23 & VC B7 & VC C3 & VC A3 \\
\hline
\end{tabular}
\end{minipage}\begin{minipage}{0.48\textwidth}
\centering
\begin{tabular}{|c|c|c|c|}
\hline
\textbf{HSA} & \textbf{Damage 1} & \textbf{Damage 2} & \textbf{Damage 3} \\
\hline
24 & VC C7 & - & - \\
25 & VC A0 & - & - \\
26 & VC B0 & - & - \\
27 & VC C0 & - & - \\
28 & HC A0 & HC A1 & - \\
29 & VC A2 & - & - \\
30 & VC A7 & - & - \\
31 & VC B4 & - & - \\
32 & VC B7 & - & - \\
33 & VC C3 & - & - \\
34 & VC C4 & - & - \\
\rowcolor{gray!20} 35 & VC C6 & - & - \\
36 & VC C7 & - & - \\
37 & B A1 & - & - \\
38 & B A2 & - & - \\
39 & B A3 & - & - \\
40 & B B1 & - & - \\
41 & B B2 & - & - \\
42 & B B3 & - & - \\
43 & B C1 & - & - \\
44 & B C2 & - & - \\
45 & B C3 & - & - \\
\hline
\end{tabular}
\end{minipage}
\label{tbl: all hsas}
\end{table*}

Photos of all HSAs used in this work are detailed in Figure \ref{fig:all hsas} and their damage and repair configurations are described in Table \ref{tbl: all hsas}.

\end{appendix}

\end{document}